\documentclass[10pt,journal,compsoc]{IEEEtran}
\usepackage{graphicx}
\usepackage{amsmath}
\usepackage{amssymb}
\usepackage{booktabs}
\usepackage{multirow}
\usepackage{caption}
\usepackage{colortbl}

\usepackage{diagbox}
\usepackage[pagebackref,breaklinks,colorlinks]{hyperref}
\usepackage{floatrow}

\def\eg{\emph{e.g.}} 
\def\ie{\emph{i.e.}} \def\Ie{\emph{I.e.}}

\def\etc{\emph{etc.}}

\usepackage[capitalize]{cleveref}
\crefname{section}{Sec.}{Secs.}
\Crefname{section}{Section}{Sections}
\Crefname{table}{Table}{Tables}
\crefname{table}{Tab.}{Tabs.}
\newcommand{\dwn}[1]{$^{\downarrow #1}$}

\def\thename{GKT}

\ifCLASSOPTIONcompsoc
  \usepackage[nocompress]{cite}
\else
  \usepackage{cite}
\fi
\ifCLASSINFOpdf
\else
\fi
\begin{document}
%
\title{Efficient and Robust  2D-to-BEV  Representation \\ Learning 
via Geometry-guided \\Kernel Transformer}

%
%
%

\author{Shaoyu Chen$^\ast$, Tianheng Cheng$^\ast$, Xinggang Wang$^\dag$, Wenming Meng, Qian Zhang, Wenyu Liu

\IEEEcompsocitemizethanks{
    \IEEEcompsocthanksitem  This work is still in progress.
    \IEEEcompsocthanksitem Shaoyu Chen, Tianheng Cheng, Xinggang Wang and Wenyu Liu are with Huazhong University of Science and Technology.
    Wenming Meng and Qian Zhang are with Horizon Robotics.
    This work is done when Shaoyu Chen and Tianheng Cheng are interns at Horizon Robotics.
    \IEEEcompsocthanksitem  $^\ast$ Equal contribution.
    \IEEEcompsocthanksitem  $^\dag$ Corresponding author.
}
}

\IEEEtitleabstractindextext{%
\begin{abstract}

Learning Bird's Eye View (BEV) representation from surrounding-view cameras is of great importance for autonomous driving. In this work, we propose a Geometry-guided Kernel Transformer (\thename), a novel 2D-to-BEV representation learning mechanism. \thename\
leverages the geometric priors to guide the transformer to focus on discriminative regions, and unfolds kernel features to generate BEV representation. For fast inference, we further introduce a look-up table (LUT) indexing method to  get rid of the camera's calibrated parameters at runtime.
\thename\ can run at $72.3$ FPS on 3090 GPU / $45.6$ FPS on 2080ti GPU and is robust to the camera deviation and the predefined BEV height.
And \thename\ achieves the state-of-the-art real-time segmentation results, i.e., 38.0 mIoU (100m$\times$100m perception range at a 0.5m resolution) on the nuScenes val set.
Given the efficiency, effectiveness and robustness,
\thename{} has great practical values in autopilot scenarios, 
especially for real-time running systems.
Code and models will be available at \url{https://github.com/hustvl/\thename}.
\end{abstract}

}

\maketitle

\IEEEdisplaynontitleabstractindextext

%
\IEEEpeerreviewmaketitle

\IEEEraisesectionheading{\section{Introduction}\label{sec:introduction}}

Surrounding-view perception based on Bird's Eye View (BEV) representation is  a cutting-edge paradigm in autonomous driving.
For multi-view camera system, how to transform 2D image representation to BEV representation is a challenging problem.
According to whether geometric information is explicitly leveraged for feature transformation,  previous methods can be divided into two categories, \ie,  geometry-based pointwise transformation and geometry-free global transformation.

\begin{figure*}
    \centering
    \includegraphics[width=\linewidth]{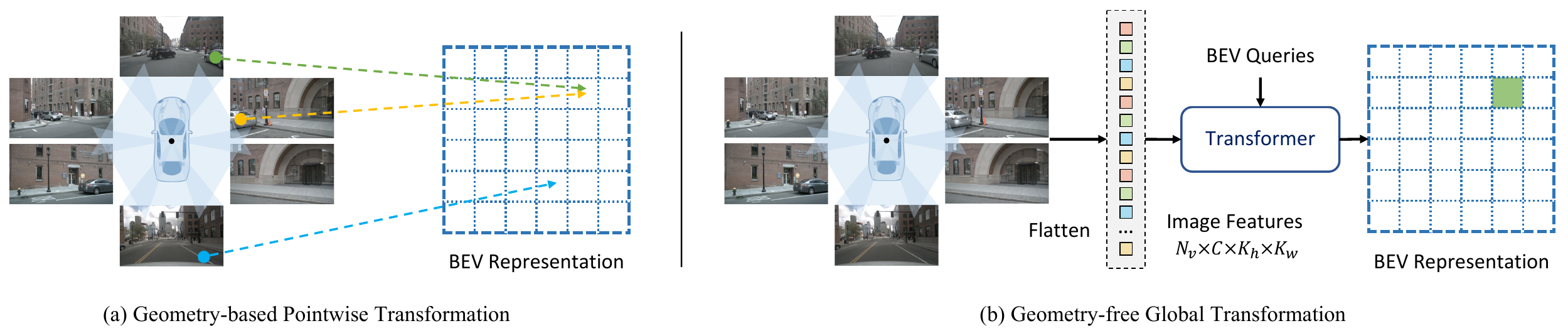}
    \caption{(a) Geometry-based pointwise transformation leverages camera's calibrated parameters (instrinsics and extrinsics) to determine the correspondence (one to one or one to many) between 2D positions and BEV grids. (b) Geometry-free global transformation considers the full correlation between image and BEV. Each BEV grid interacts with all image pixels.}
    \label{fig:pointwise-global}
\end{figure*}

\noindent\textbf{Geometry-based Pointwise Transformation} \ 
As illustrated in Fig.~\ref{fig:pointwise-global}(a), pointwise transformation methods~\cite{fiery,lss,bevformer,oft,pon,bevdet}
leverage camera's calibrated parameters (instrinsics and extrinsics) to determine the correspondence (one-to-one or one-to-many) between 2D positions and BEV grids.
With the correspondence available, 2D features are projected to 3D space and form BEV representation.

However, pointwise transformation relies too much on the calibrated parameters. For a running autopilot system, because of the  complicated external environment, cameras might deviate from the calibrated position at runtime,
which  makes the  2D-to-BEV correspondence unstable and brings system error to the BEV representation.
Besides, pointwise transformation usually requires complicated and time-consuming 2D-3D mapping operations, \eg, 
predicting depth probability distribution over pixels, broadcasting pixel features along the ray to BEV space, and high precision calculation about camera's parameters. These operations are inefficient and hard to be optimized, limiting the real-time applications.

\noindent\textbf{Geometry-free Global Transformation} \ 
Global transformation methods~\cite{cvt} consider the full correlation between image and BEV.
As shown in  Fig.~\ref{fig:pointwise-global}(b),  multi-view image features are flattened and each BEV grid interacts with all image pixels. 
Global transformation does not rely on the  geometric priors in 2D-to-BEV projection. Thus, it's insensitive to the camera deviation. 

But it also raise problems.
1)  The computational budget of global transformation is proportional to the number of image pixels. There exists sharp contradiction between resolution and efficiency.
2) Without geometric priors as guidance, the model
has to globally dig out discriminative information from all views, which makes convergence harder.

In this work,  targeting at efficient and robust BEV representation learning, we propose a new 2D-to-BEV transformation mechanism, named Geometry-guided Kernel Transformer (\thename{} for short). 
With  coarse  cameras' parameters, we roughly project BEV positions to get  prior 2D positions in multi-view and multi-scale feature maps.
Then, we unfold $K_h \times K_w$  kernel features around the prior positions, and make BEV queries interact with corresponding  unfolded features to generate BEV representation. Further, we introduce LUT indexing to get rid of camera’s parameters at runtime.

\thename{} is of high robustness at runtime.
Compared with pointwise transformation, \thename\ only takes camera's parameters as guidance but not rely too much on them. 
When camera deviates, correspondingly the kernel regions shift but can still cover the targets.
Transformer is permutation-invariant and the attention weights for the kernel regions are dynamically generated according to the deviation.
Thus, \thename{} can always focus on the targets, insensitive to the deviation of camera. 

\thename{} is of high efficiency. 
With the proposed LUT indexing, at runtime we get rid of 2D-3D mapping operations required by pointwise transformation, making the forward process compact and fast.
And compared with global transformation, \thename\ only focuses on geometry-guided kernel regions, avoiding global interaction. \thename\ requires less computation and convergences faster.

Consequently, \thename\  well balances  between pointwise and global transformation, leading to efficient and robust 2D-to-BEV representation learning.
We  validate \thename\ on nuScenes map-view segmentation. \thename\ is promisingly efficient,  running at  \textbf{72.3} FPS  on 3090 GPU / \textbf{45.6} FPS on  2080ti GPU, much faster than all existing methods.
And \thename\ achieves \textbf{38.0} mIoU, which is SOTA among all real-time methods.
We will extend \thename{} to other BEV-based tasks in the near future.

\section{Method} 
\subsection{Geometry-guided Kernel Transformer} 
\begin{figure*}
    \centering
    \includegraphics[width=\linewidth]{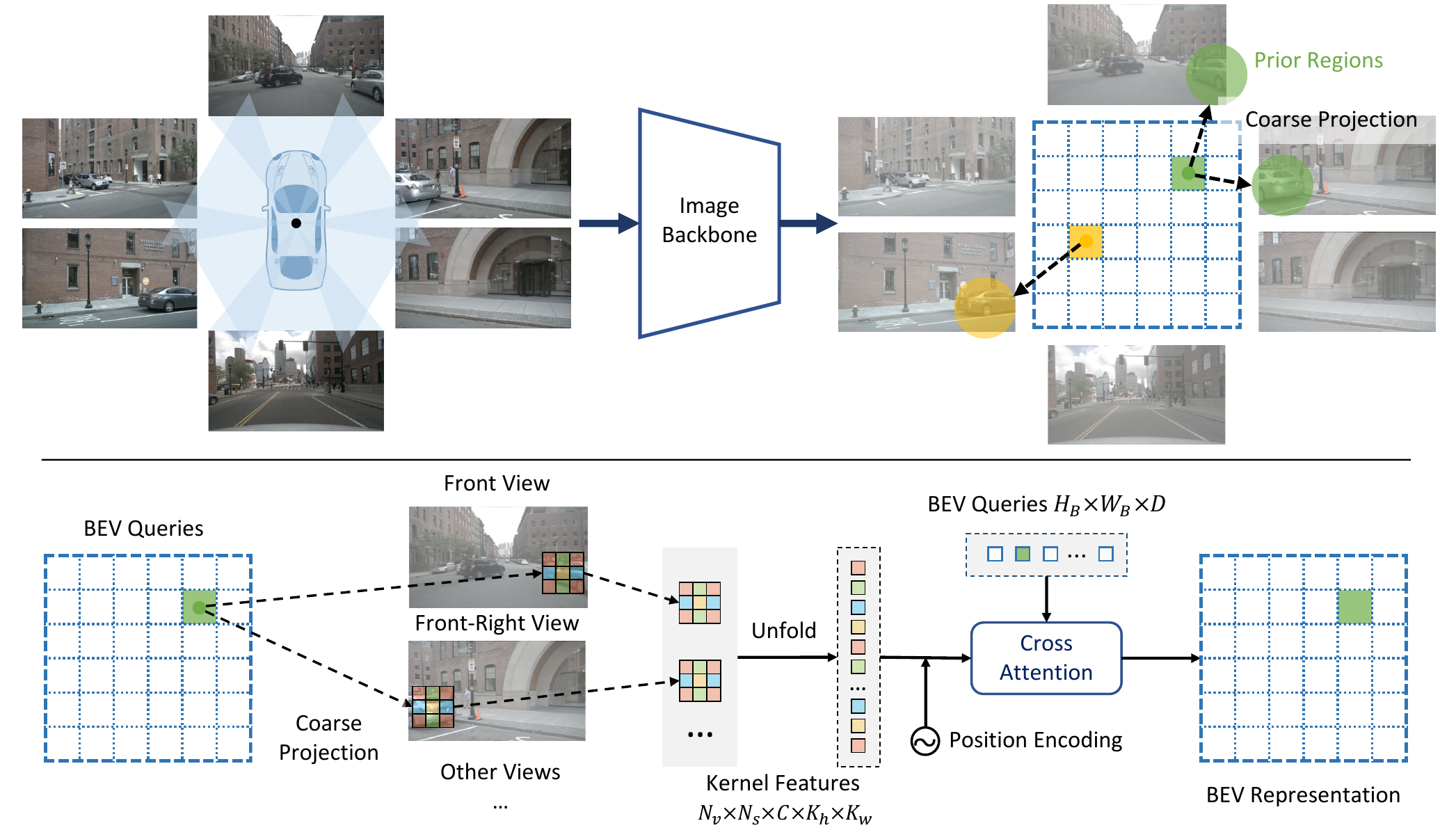}
    \caption{Illustration of \thename. Top:  geometric information is leveraged to guide the transformer to focus on prior regions in multi-view images. Bottom: we unfold kernel features of the prior regions and make them interact with BEV queries to generate BEV representation.}
    \label{fig:GKT-main}
\end{figure*}
The framework of the proposed \thename\ is presented in Fig~\ref{fig:GKT-main}.
Shared CNN backbone  extracts multi-scale multi-view features $\mathcal{F}_{\text{img}}=\{{F}^s_v\}$ from surround-view images $\mathcal{I}=\{{I}_v\}$.
The BEV space is evenly divided into grids. Each BEV grid corresponds to a 3D coordinate $P_i = (x_i, y_i, z)$ and a learnable query embedding $q_i$.
$z$ is the predefined height of BEV plane shared by all queries.   \thename\ is insensitive to the value of $z$, which is further discussed in Sec.~\ref{sec:bev_height}.

We leverage the geometric priors to guide the transformer to focus on discriminative regions.
With  camera's parameters,
we roughly project each BEV grid $P_i$ to a set of float 2D pixel coordinates $\{Q_i^{sv}\}$ (for different views and scales) and then round them to integer coordinates $\{\bar{Q}_i^{sv}\}$, \ie,
\begin{equation}
\begin{aligned}
    & Q_i^{sv} = \mathbf{K}^{sv} \cdot \mathbf{Rt}^{sv} \cdot P_i^{sv},\\
    & \bar{Q}_i^{sv} = round(Q_i^{sv})
\end{aligned}
\label{eq:projection}
\end{equation}
We unfold  $K_h \times K_w$ kernel regions around the prior positions $\{\bar{Q}_i^{sv}\}$.
It's worth noting that if the kernel regions exceed the image boundary the exceeding part is set to zero.
Each BEV query $q_i$ interacts with the corresponding unfolded kernel features $F \in  \mathbb{R}^{N_{view} \times N_{scale} \times C \times K_h \times K_w}$
and  generates BEV presentation.
Heads for various tasks (\eg, detection, segmentation, motion planning)  can be performed on the BEV presentation.

\subsection{Robustness to Camera Deviation} 

For a running autopilot system, the external environment is complicated. Camera  will deviate from its calibrated position.
\thename{} is robust to the camera deviation. To validate this,
we  simulate  the  deviation of camera in real scenarios.
Specifically,  we decompose the deviation to rotation deviation  $\mathbf{R}_{devi}$  and 
translation deviation  $\mathbf{T}_{devi}$ and 
 add random noise to all the $x,y,z$ dimensions. 

The translation deviation  $\mathbf{R}_{devi}$ is formulated as,
\begin{equation}
\begin{aligned}
  \mathbf{T}_{devi} &= 
  \begin{bmatrix}
    1 & 0 & 0 & \Delta x\\
    0 & 1 & 0 & \Delta y\\
    0 & 0 & 1 & \Delta z\\
    0 & 0 & 0 & 1 \\
  \end{bmatrix}
\end{aligned}
\end{equation}

The rotation deviation  $\mathbf{R}_{devi}$ is formulated as,
\begin{equation}
    \begin{aligned}
          \mathbf{R}_{devi} =  R_{\theta_x}  \cdot  R_{\theta_y} \cdot R_{\theta_z} \\
    \end{aligned}
\end{equation}

\begin{equation}
\begin{aligned}
  R_{\theta_x} = \begin{bmatrix}
    1 & 0 & 0 & 0\\
    0 & cos(\theta_x) & sin(\theta_x) & 0\\
    0 &  -sin(\theta_x) & cos(\theta_x)  & 0\\
    0 & 0 & 0 & 1 \\
  \end{bmatrix}
  \\
  R_{\theta_y} = 
    \begin{bmatrix}
    cos(\theta_y)  &  & -sin(\theta_y) & 0\\
    0 & 1 & 0 & 0\\
    sin(\theta_y) & 0 & cos(\theta_y)  & 0\\
    0 & 0 & 0 & 1 \\
  \end{bmatrix}
  \\ 
  R_{\theta_z} =
    \begin{bmatrix}
    cos(\theta_z) & sin(\theta_z) & 0 & 0\\
    -sin(\theta_z) & cos(\theta_z) & 0 & 0\\
    0 & 0 & 1 & 0\\
    0 & 0 & 0 & 1 \\
  \end{bmatrix}
\end{aligned}
\end{equation}

$\Delta x, \Delta y, \Delta z, \theta_x, \theta_y, \theta_z$  are all
 random variables subject to normal distribution.
 \Ie,
\begin{equation}
\begin{aligned}
\Delta x, \Delta y, \Delta z \sim \mathcal{N}(0,{\sigma_2}^2)\\
\theta_x, \theta_y, \theta_z  \sim \mathcal{N}(0,{\sigma_1}^2)
\end{aligned}
\end{equation}

$\theta_x, \theta_y, \theta_z$
correspond to the rotation noise and $\Delta x, \Delta y, \Delta z$ correspond to the translation noise,
respectively relative to the the  $x, y, z$  axes of camera coordinate system.

We add random noise to all the $x,y,z$ dimensions and all cameras.
With deviation noise introduced, Eq.~\ref{eq:projection} becomes
\begin{equation}
\begin{aligned}
    & Q_i^{sv} = \mathbf{K}^{sv} \cdot \mathbf{R}_{devi} \cdot \mathbf{T}_{devi}  \cdot  \mathbf{Rt}^{sv} \cdot P_i^{sv},\\
    & \bar{Q}_i^{sv} = round(Q_i^{sv})
\end{aligned}
\end{equation}

Experiments about the camera deviation are presented in Sec.~\ref{sec:camera_deviation}.
\thename\ is  robust to the deviation. 
 When camera deviates, the prior regions shift but can still
cover the targets. 
And the rounding operation in Eq~\ref{eq:projection} is anti-noise. When the camera's parameters slightly change, after rounding the coordinates  keep the same.
Besides, transformer is permutation-invariant and the attention weights for the kernel regions are dynamic according to the deviation. 

\subsection{BEV-to-2D LUT Indexing} 
We further introduce BEV-to-2D LUT Indexing to speed up \thename.
The kernel regions for each BEV grid is fixed and can be pre-computed offline.
Before runtime, we construct a LUT (look-up table) which caches the correspondence between indices of BEV queries and indices of image pixels.
At runtime, we get corresponding pixel indices for each BEV query from the LUT, and efficiently fetch kernel features through indexing. 

With LUT Indexing, \thename{} gets rid of high precision calculation about camera's parameters  and achieves higher FPS. 
In Sec.~\ref{sec:implementation}, we compare other implementation manners with  LUT Indexing to validate the efficiency.

\subsection{Configuration of Kernel} 
For \thename, the configuration of kernel is flexible. 
We can adjust the kernel size to balance between the receptive field and computation cost. 
And the layout of kernel is feasible  (cross shape kernel, dilated kernel, \etc). Since LUT indexing is adopted to fetch kernel features, the efficiency of \thename\ is not influenced by the layout of kernel.

\section{Experiments} 
In this section, we mainly evaluate the proposed \thename{} on nuScenes map-view segmentation and conduct extensive experiments to investigate \thename{}, including the convergence speed and robustness to the camera deviation.

\subsection{Dataset}
The nuScenes dataset~\cite{nuScenes} contains 1,000 driving sequences with 700, 150, and 150 for training, validation, and testing, respectively.
Each sequence lasts 20 nearly seconds and contains 6 surrounding-view images around the ego-vehicle per frame.
We resize the 6-camera images into $224\times480$ for both training and validation.
All models are trained on nuScenes \textit{train} set and evaluated on the \textit{val} set.

\subsection{Implementation Details}
We adopt CVT~\cite{cvt} as the basic implementation for BEV map-view segmentation.
Following ~\cite{cvt}, we feed the images into an EfficientNet-B4~\cite{efficientTanL19} to extract multi-scale image features.
We randomly initialize the $25\times25$ BEV feature maps as queries and adopt the proposed \thename{} to transform image features to BEV features.
Then three convolutional blocks with upsampling layers are employed to upsample BEV features to $200\times200$ for map-view segmentation.

All models are trained on 4 NVIDIA GPUs with 4 samples per GPU.
We adopt the training schedule from \cite{cvt} and train the \thename{} with 30k iterations and AdamW optimizer, which takes nearly 3 hours for training.
Following \cite{cvt}, we adopt two evaluation settings, \ie, \textbf{setting 1} and \textbf{setting 2}, which evaluate the segmentation results with 100m$\times$50m perception range at a 0.25m resolution and 100m$\times$100m perception range at a 0.5m resolution, respectively.
Unless specified, we adopt \textbf{setting 2} as the default evaluation setting.

\subsection{Main Results} 
In Tab.~\ref{tab:main_results}, we compare \thename\ with other BEV-based methods on vehicle map-view segmentation with two evaluation settings.
Specifically, we adopt a $1\times7$ convolution for capturing the horizontal context and then apply an efficient $7\times1$ kernel for 2D-to-BEV transformation in \thename{}.
FIERY~\cite{fiery} and BEVFormer~\cite{bevformer} achieve high performance but are time-consuming, far away from real-time applications.  
\thename\ can run at $45.6$ FPS on 2080Ti with $41.4$ mIoU for \textbf{setting 1} $38.0$ mIoU for \textbf{setting 2},
achieving the highest efficiency and performance among all existing real-time methods.

\begin{table}[]
    \centering
    \caption{\textbf{Vehicle map-view segmentation on nuScenes.} FPS is measured on 2080Ti GPU, except $^*$ denotes measured on V100. $^\dagger$ denotes our reproduced results, which are higher than the ones reported in the original paper.}
    \label{tab:main_results}
    \begin{tabular}{l|cccc}
    \toprule
    Method & Setting 1 & Setting 2 & FPS & Params \\
    
    \hline
    VPN~\cite{vpn} & 25.5 & - & - & - \\
    STA~\cite{sta} & 36.0 & - & - & - \\
    FIERY~\cite{fiery} & 37.7 & 35.8 & - & - \\
    FIERY$^\dagger$~\cite{fiery} & 42.7 & 39.8 & 8.0 & 7.4M \\
    BEVFormer~\cite{bevformer} & - & 43.2 & 1.7$^*$ & 68.1M  \\
    \hline
    PON~\cite{pon} & 24.7 & - & 30.0 & -\\
    Lift-Splat~\cite{lss} & - &32.1 & 25.0 & 14.0M \\
    CVT~\cite{cvt} & 37.2 & 36.0 & 35.0 & 1.1M \\
    CVT$^\dagger$~\cite{cvt} & 39.3 & 37.2 & 34.1 & 1.1M \\
    \thename\ & 41.4 & 38.0 & 45.6 & 1.2M \\
    \bottomrule
    \end{tabular}
\end{table}




\subsection{Robustness to Camera Deviation} 
\label{sec:camera_deviation}
To validate the robustness, we traverse the validation sets  (with $6019$ samples) and randomly generate noise for each sample. We respectively add translation and rotation deviation of different degrees.
Note that we add noise to all cameras and all coordinates. 
And the noise is subject to normal distribution. There exists extremely large deviation in some samples, which affect the performance a lot.
As shown in Tab.~\ref{tab:translation}  and Tab.~\ref{tab:rotation},  when the standard deviation of  $\Delta_x, \Delta_y, \Delta_z$ is  $0.5m$ or the standard deviation of $\theta_x, \theta_y, \theta_z$ is  $0.02rad$,
\thename{}  still keeps comparable performance.

We observe that $5\times5$ kernel is more robust to the deviation than $3\times3$ kernel.
It proves larger kernel size corresponds to stronger robustness.
And $7\times3$ kernel is more robust to the deviation than $5\times5$ kernel. 
$7$ ($k_h$) corresponds to the vertical dimension of the 3D space. For each BEV grid, $x, y$ is fixed but $z$ is predefined. There exists more uncertainty in vertical dimension. Thus, adopting to a larger $k_h$ is better.

\begin{table}[h]
    \centering
    \caption{\textbf{Robustness to the translation deviation of camera.} The metric is mIoU. $\sigma_1$ is the standard deviation of  $\Delta_x, \Delta_y, \Delta_z$.} 
    \label{tab:translation}
    \begin{tabular}{l|cccccc}
    \toprule
    \multirow{2}{*}{Kernel} & \multicolumn{5}{c}{ $\sigma_1 (m)$} \\
    \cline{2-6} & 0 & 0.05 & 0.1 & 0.5 & 1.0   \\
    \hline
    $3\times3$ & 36.5 & 36.5\dwn{0.0} & 36.3\dwn{0.2} & 33.1\dwn{3.4} & 27.3\dwn{9.2} \\
    $5\times5$ & 36.6 & 36.5\dwn{0.1} & 36.4\dwn{0.2} & 33.9\dwn{2.7} & 28.8\dwn{7.8} \\
    $7\times3$ & 37.3 & 37.3\dwn{0.0} & 37.1\dwn{0.2} & 34.9\dwn{2.4} & 30.5\dwn{6.8} \\
    \bottomrule
    \end{tabular}
\end{table}

\begin{table}[h]
    \centering
    \caption{\textbf{Robustness to the rotation deviation of camera.} The metric is mIoU. $\sigma_2$ is the standard deviation of  $\theta_x, \theta_y, \theta_z$.}
    \label{tab:rotation}
    \begin{tabular}{l|cccccc}
    \toprule

    \multirow{2}{*}{Kernel} & \multicolumn{5}{c}{$\sigma_2 (rad)$} \\
    \cline{2-6} & 0 & 0.005 & 0.01 & 0.02 & 0.05  \\
    \hline
    $3\times3$ & 36.5 & 36.2\dwn{0.3} & 35.5\dwn{1.0} & 33.6\dwn{2.9} & 25.6\dwn{10.9} \\
    $5\times5$ & 36.6 & 36.3\dwn{0.3} & 35.7\dwn{0.9} & 34.1\dwn{2.5} & 27.4\dwn{9.2} \\
    $7\times3$ & 37.3 & 37.0\dwn{0.3} & 36.5\dwn{0.8} & 34.9\dwn{2.4} & 28.4\dwn{8.9} \\
    \bottomrule
    \end{tabular}
\end{table}

\subsection{Robustness to BEV Height}
\label{sec:bev_height}
In Tab.~\ref{tab:bev_height}, we  ablate on the predefined height $z$ of BEV plane.  When $z$ varies from $-1.0$ to $2.0$,  the performance fluctuates little (smaller than $0.4$ mIoU).  
The predefined height $z$ affects the BEV-to-2D projection. But \thename\ only requires coarse projection and thus is quite insensitive to the value of $z$. The experiments prove the robustness of \thename. 

\begin{table}[h]
    \centering
    \caption{\textbf{Robustness to the predefined height $z$ of BEV plane.} The metric is mIoU. The performance of \thename{} fluctuates little.}
    \label{tab:bev_height}
    \begin{tabular}{l|cccc}
    \toprule
    Kernel & z=-1.0 & z=0.0 & z=1.0 & z=2.0 \\
    \hline
    $3\times3$ & 36.5 & 36.4 & 36.4 & 36.6 \\
    $5\times5$ & 36.4 & 36.8 & 36.7 & 36.5 \\
    \bottomrule
    \end{tabular}
\end{table}

\subsection{Convergence Speed} 
\begin{figure}[h]
    \centering
    \includegraphics[width=\linewidth]{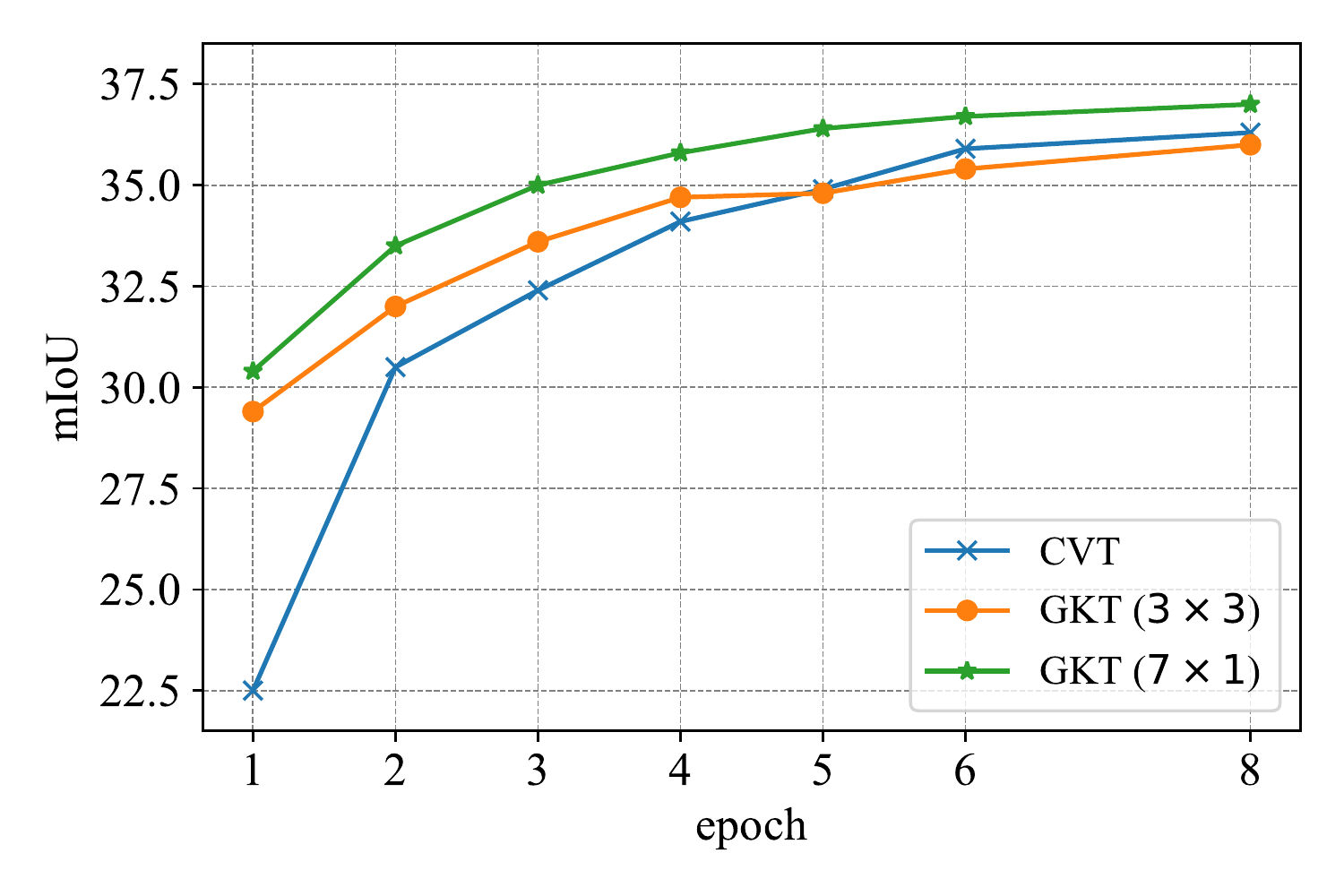}
    \caption{\textbf{Comparison of the convergence speed.}
    We train  \thename\ and CVT~\cite{cvt} with different training schedules.
    For only 1-epoch training schedule, the gap of mIoU is obvious.  \thename\ converges much faster than CVT~\cite{cvt}.}
    \label{fig:convergence}
\end{figure}

Fig.~\ref{fig:convergence} compares the convergence speed between \thename{} and CVT~\cite{cvt}.
We evaluate the models with different training schedules, \ie, from 1-epoch to 8-epoch  schedule.
Based on the geometric priors, the proposed \thename{} achieves faster convergence speed than CVT and obtains much better results with only 1-epoch training.

\subsection{Implementation of \thename{}}
\label{sec:implementation}
The proposed 2D-to-BEV transformation can be implemented in different manners. 

1) Im2col: we first adopt \texttt{im2col} operation to unfold image features to the column formats.
Each column corresponds to a kernel region.
Then we select the corresponding column for each BEV query. Im2col is straightforward but results in large memory consumption.

2) Grid Sample: we adopt \texttt{grid\_sample} operation to sample  features of all pixels in the kernel regions, and concat them together.

3) LUT Indexing: to further accelerate the inference of transformation, we pre-compute the correspondence between  indices of BEV grids and indices of image pixels and then building a look-up table.

Tab.~\ref{tab:ablation_gkt_impl} compares the inference speeds (FPS) of different implementation manners (kernel is $3\times3$). LUT Indexing achieves the best inference speed as expected.


\begin{table}[h]
    \centering
    \caption{\textbf{Comparison of different implementation manners.} We adopt different implementations of the proposed \thename{} and evaluate the inference speed on one NVIDIA 2080Ti.}
    \label{tab:ablation_gkt_impl}
    \begin{tabular}{c|ccc}
    \toprule
    Imp. & Im2col & Grid Sample & LUT Indexing \\
    \hline
    FPS & 38.8 & 42.3 & 43.4 \\
    \bottomrule
    \end{tabular}
\end{table}

\section{Related Work}
\subsection{BEV Representation Learning}
Bird's Eye View is a promising representation in autonomous driving.
Recent works study how to transform image representation to BEV representation.
Lift-Splat~\cite{lss}, FIERY~\cite{fiery} and BEVDet~\cite{bevdet} lift camera features to 3D by predicting a depth probability distribution over pixels and using known camera intrinsics and extrinsics.
OFT~\cite{oft} proposes orthographic feature transformation.
BEVFormer~\cite{bevformer}  predicts height offsets of BEV anchor points and then use camera's parameters to fix the 2D-to-3D correspondence.
PON~\cite{pon} condenses the image features along the vertical dimension and then adopts dense transformer layer to expand features along the depth axis. The above methods all require precise calibrated camera's parameters for 2D-3D transformation. Differently,
CVT~\cite{cvt} globally aggregates information from all image features  through a series of cross attention layers to generate BEV  representation, without explicitly adopting parameters. 
\thename\ is also robust to camera's parameters but achieves better efficiency by leveraging  geometric priors.

\subsection{Perception and Planning based on BEV}
Previous works study the perception and planning based on Bird's Eye View representation.
FIERY~\cite{fiery} predicts temporally consistent future instance segmentation and motion in bird’s-eye view.
\cite{lss, pon} conducts map segmentation based on BEV.
\cite{bevdet, oft} adopts BEV representation for 3D object detection.
BEVFormer~\cite{bevformer} builds up detection and segmentation head on BEV feature maps for multi-task learning.
\thename\ provides robust BEV representation. Various perception and planning tasks can be based on \thename.

\section{Conclusion} 
We present \thename\ for 2D-to-BEV representation learning.
\thename\ has high efficiency and robustness, both of which are crucial characteristics for a real-time running system, especially for autopilot.
We validate \thename\ on map-view segmentation. In the near future, we will extend it to other BEV-based tasks, like detection and motion planning.



\ifCLASSOPTIONcaptionsoff
  \newpage
\fi


%

{\small
\bibliographystyle{IEEEtran}
\bibliography{egbib}
}

%








\end{document}